\documentclass{article}

\PassOptionsToPackage{numbers, compress}{natbib}
\usepackage[preprint]{arxiv}

\usepackage[utf8]{inputenc} % allow utf-8 input
\usepackage[T1]{fontenc}    % use 8-bit T1 fonts
\usepackage{hyperref}       % hyperlinks
\usepackage{url}            % simple URL typesetting
\usepackage{booktabs}       % professional-quality tables
\usepackage{amsfonts}       % blackboard math symbols
\usepackage{nicefrac}       % compact symbols for 1/2, etc.
\usepackage{microtype}      % microtypography
\usepackage{xcolor}         % colors
\usepackage{graphicx}
\usepackage{amsmath}
\usepackage{amssymb}
\usepackage{cleveref}
\usepackage{subcaption}
\usepackage{authblk}

\title{~Developing hybrid mechanistic and data-driven personalized prediction models for platelet dynamics~}

\author{
  Marie Steinacker$^{1,2}$ \thanks{Corresponding author: \texttt{steinacker@informatik.uni-leipzig.de}} \quad
  Yuri Kheifetz$^2$ \quad
  Markus Scholz$^2$ \\
 $^1$Center for Scalable Data Analytics and Artificial Intelligence (ScaDS.AI) Dresden/Leipzig,\\
  Universität Leipzig, Leipzig, Germany\\
  $^2$Institute for Medical Informatics, Statistics and Epidemiology
  (IMISE),
  Universität Leipzig, Leipzig, Germany

}

\begin{document}

\maketitle

\begin{abstract}
 Hematotoxicity, drug-induced damage to the blood-forming system, is a frequent side effect of cytotoxic chemotherapy and poses a significant challenge in clinical practice due to its high inter-patient variability and limited predictability. Current mechanistic models often struggle to accurately forecast outcomes for patients with irregular or atypical trajectories. In this study, we develop and compare hybrid mechanistic and data-driven approaches for individualized time series modeling of platelet counts during chemotherapy. We consider hybrid models that combine mechanistic models with neural networks, known as universal differential equations. As a purely data-driven alternative, we utilize a nonlinear autoregressive exogenous model using gated recurrent units as the underlying architecture. These models are evaluated across a range of real patient scenarios, varying in data availability and sparsity, to assess predictive performance. Our findings demonstrate that data-driven methods, when provided with sufficient data, significantly improve prediction accuracy, particularly for high-risk patients with irregular platelet dynamics. This highlights the potential of data-driven approaches in enhancing clinical decision-making. In contrast, hybrid and mechanistic models are superior in scenarios with limited or sparse data. The proposed modeling and comparison framework is generalizable and could be extended to predict other treatment-related toxicities, offering broad applicability in personalized medicine.
\end{abstract}

\section{Introduction}
\label{introduction}
Cytotoxic chemotherapy induces dose-limiting hematotoxic side effects such as neutropenia (lack of white cells) or thrombocytopenia (lack of platelets) \cite{pfreundschuh_two-weekly_2004-1, pfreundschuh_two-weekly_2004, crawford_chemotherapyinduced_2004}. A central challenge in precision medicine is to predict an individual's toxicity risk in subsequent treatment cycles to enable adaptive, personalized therapy planning. To address this challenge, several statistical \cite{wunderlich_practicability_2003}, mechanistic \cite{friberg_model_2002, friberg_mechanistic_2003, mangas-sanjuan_semimechanistic_2015, henrich_semimechanistic_2017} and comprehensive mechanistic models \cite{scholz_biomathematical_2010, kheifetz_modeling_2019, kheifetz_individual_2021} of bone marrow hematopoiesis under chemotherapy were proposed, but with limited success. In particular, patients with irregular dynamics and deep blood cell nadirs were insufficiently described by existing models. We hypothesize that mechanistic models do not adequately capture inter-individual variability due to potentially incomplete or overly simplistic system representations, leading to limited flexibility and prediction performance. 

Data-driven neural network-based forecasting methods are widely applied to prediction tasks in clinical settings, e.g., for diagnosis \cite{choi_doctor_2016} or the prediction of clinical events \cite{esteban_predicting_2016}. Neural networks are powerful universal function approximators \cite{cybenko_approximation_1989}, but typically require large, high-quality data sets for good performance. In contrast, in clinical settings data for individual patients are typically sparse and noisy. Multiple attempts have been made to combine mechanistic and data-driven modeling, either by incorporating mechanistic knowledge into hybrid models \cite{rackauckas_universal_2021}, in the loss function \cite{raissi_physics-informed_2019}, or through transfer learning \cite{kleissl_simbaml_2023, steinacker_individual_2023, zabbarov_optimizing_2024}. Here, we aim to develop and compare methods for the individualized prediction of chemotherapy induced platelet toxicity based on time series forecasting. We design both hybrid and purely data-driven approaches, and evaluate their relative performance against popular mechanistic models on a clinical data set (52.8k individual measurements in total for 360 patients).

\begin{figure}[h]
    \centering
    \includegraphics[width=\textwidth]{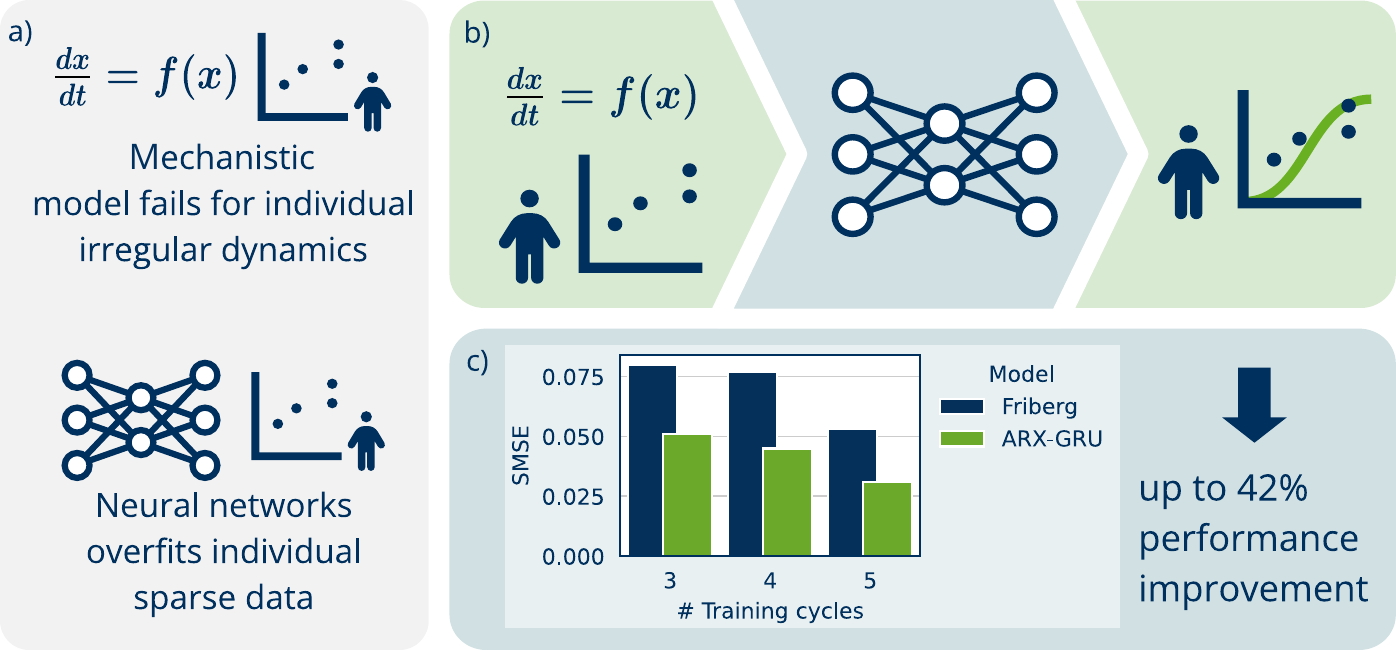}
    \caption{Combining mechanistic and data-driven models. Mechanistic models and data-driven approaches alone fail to perform well for individual sparse patient data (a). To deal with this problem, we employ transfer learning and hybrid modeling approaches, where both, individual patient dynamics and existing mechanistic knowledge is combined in model training (b). Results showed a strong advantage of the data-driven ARX-GRU approach, if sufficient data is provided for training (c).}
    \label{fig:graph_abstract}
\end{figure}

To construct hybrid models, we combine a mechanistic model, further referred to as the Friberg model \cite{friberg_model_2002}, with a Universal differential equation approach (UDE) \cite{rackauckas_universal_2021}. A similar approach has been undertaken by Martensen et al. \cite{martensen_data-driven_2024}, who replaced mechanistic feedbacks with deep nonlinear mixed effect modeling in order to extract mathematical equations for the newly learned mechanisms via symbolic regression. In contrast to this population based approach, we here pursue an individual approach building small hybrid models for individual patients. For comparison, we examine nonlinear autoregressive networks with exogenous input (NARX) as purely data-driven alternatives \cite{siegelmann_computational_1997}. These models showed decent success in prediction of platelet dynamics in the past using transfer learning \cite{steinacker_predicting_2024}, but were not yet compared with refined mechanistic models or hybrid approaches. Anticipating that there is probably no unique modeling approach which is optimal for all individual situations, we investigate how data sparsity and variability affect prediction performance. We present a graphical summary of our study in \Cref{fig:graph_abstract}.

\paragraph{Contributions.}
To address the need for individualized forecasting of platelet toxicity, we develop suitable frameworks of hybrid and data-driven models to predict individual platelet dynamics under chemotherapy. We compare their predictive performance to established mechanistic models and examine how model effectiveness varies with individual data characteristics. In particular, we analyze the impact of the number of available therapy cycles, data points used for training and treatment intensity affecting platelet variability on the importance of data-driven model components. Our results highlight the benefits of incorporating data-driven elements into individualized modeling. Ultimately, our methods may support medical professionals in developing personalized treatment strategies, with the potential to improve clinical outcomes. The code implementing our methods and learning approaches is publicly available at \href{https://github.com/earlgreymatchalatte/Hybrid_Data-Driven_Models}{GitHub}, enabling application to other data sets.

\section{Time series forecasting models}
We describe our modeling and learning approaches in the following. Details on hyperparameter selection for all models are provided in the supplement. Code to implementations is also linked in the supplement.

\subsection{Mechanistic models}
By mechanistic models we refer to differential equation models that are based on simplifying physiological assumptions of the underlying biological processes - in our case hematopoiesis and chemotherapy treatment effects thereon. The advantage of such models is that biological knowledge is explicitly represented. However, the predetermined structure of the equations constrains the solution space and limits the model’s flexibility. We consider several mechanistic models of hematopoiesis under chemotherapy as a baseline and reference for our hybrid approach.

\paragraph{Friberg model}
The Friberg model \cite{friberg_model_2002} is a simple phenomenological model of hematopoiesis under chemotherapy that is often employed to predict chemotherapy toxicity. It describes the proliferation of stem cells and their maturation into circulating blood cells. Cytotoxic chemotherapy is modeled by reduced stem cell proliferation. The following equations were proposed to describe these processes:
\begin{align}\label{eqn:friberg}
    \frac{d}{dt}P &= k_{p}  P  (1-E_{drug}) \left(\frac{C_0}{C}\right)^\gamma - k_{tr}  P,\\
    \frac{d}{dt}T_1 &= k_{tr}  P - k_{tr}  T_1,\\
    \frac{d}{dt}T_2 &= k_{tr}  T_1 - k_{tr}  T_2,\\
    \frac{d}{dt}T_3 &= k_{tr}  T_2 - k_{tr}  T_3,\\
    \frac{d}{dt}C &= k_{tr}  T_3 - k_{c}  C.
\end{align}
Maturation is modeled via three sub-compartments $T_1-T_3$ with first order transitions. The strength of the feedback of circulating cells $C$ on stem cell proliferation is described by parameter $\gamma$. Individual steady states are defined by $C_0$. To reduce parameters, all rate parameters are set equal ($k_p = k_c = k_{tr}$) \cite{friberg_model_2002}. Chemotherapy is modeled by an effect function $E_{drug}$ depending on the applied cytotoxic drugs and their concentrations. We assume a simple step-function, which is non-zero at days of treatment and proportional to the drug dosage.

The Friberg model is limited in its capability to describe hematopoietic long-term recovery dynamics or cumulative toxicity after chemotherapy. It has restricted dynamics which typically consist of an initial decline followed by a recovery. 

\paragraph{Mechanistic refinements of the Friberg model}
Several improvements of the Friberg model were proposed in the literature, e.g. to better explain long-term cumulative toxicity. Here, we examine the models developed by Mangas-Sanjuan et al. \cite{mangas-sanjuan_semimechanistic_2015} (MS) and Henrich et al. \cite{henrich_semimechanistic_2017}. Both models introduce additional stem cell compartment(s) to model the cumulative effect of multiple cytotoxic drug applications. We also devise a slightly modified version of the Mangas-Sanjuan model (MS-rev.), to align more closely with the current biological knowledge. Details and model equations are provided in the supplement.

\subsection{Universal differential equations}
Universal differential equations (UDEs) are a framework combining mechanistic modeling with neural network approaches to establish hybrid models combining natural and artificial intelligence \cite{rackauckas_universal_2021}. The basic idea is that parts of the differential equations of a mechanistic model are replaced or augmented with neural network components to capture dynamics that are insufficiently explained by the mechanistic part of the model. This allows a phenomenologic, data driven approach to incorporate mechanistically poorly understood biological processes into prediction models. Due to their mechanistic foundation, these hybrid approaches have better interpretability compared to pure neural network models. Using neural networks with differential equation was originally proposed by Chen et al. \cite{chen_neural_2018} and called neural ODEs. UDEs are a generalization of neural ODEs.

Perspectives to employ UDEs in biological research were discussed in \cite{rackauckas_universal_2021}. To avoid non-biological solutions, the concept of "non-negative UDEs" (n-UDE) was introduced \cite{philipps_non-negative_2024}. We adopt this concept by developing a n-UDE augmentation of the Friberg model as explained in the following. 

\paragraph{n-UDE expansion of the Friberg model}
We aim to describe individual time courses of platelets with n-UDEs using the Friberg model as basis. Our augmentation approach targets the feedback mechanism between circulating cells and stem cells because this mechanism is poorly understood. 

We examine two different versions: (1) ``UDE-add'', where we simply add a neural network (NN) to the feedback term in \Cref{eqn:friberg} and (2) ``UDE-rep'' where we replace the entire feedback term with a neural network. In detail, for the first case \Cref{eqn:friberg} is replaced by 
\begin{equation}
    \frac{d}{dt}P = k_{p}  P (1-E_{drug}) \left(\frac{C_0}{C}\right)^\gamma - k_{tr} P + \tanh(a P)\text{NN}(P, C, E_{drug}), 
\end{equation}
and in the second case, we have 
\begin{equation}
    \frac{d}{dt}P = \tanh(a P) \text{NN}(P, C, E_{drug}) (1-E_{drug}) - k_{tr}  P.
\end{equation}
Non-negativity is assured by including the $\tanh$-term with a small factor $a$, as described in  \cite{philipps_non-negative_2024}. We fixed $a$ to 0.005 by hyperparameter optimization.

\subsection{NARX networks} 
As a completely data-driven alternative, we consider Nonlinear Autoregressive Networks with Exogenous Inputs (NARX). NARX networks are a class of recurrent neural networks (RNNs) that model dynamic systems by incorporating both past outputs and external input signals to predict future values \cite{siegelmann_computational_1997}. In our case, the signal corresponds to the platelet counts while the inputs represent the chemotherapy treatment. In previous work, NARX networks with gated recurrent units (ARX-GRU) were identified as a suitable approach to our problem \cite{steinacker_predicting_2024}. We compare our UDE hybrid frameworks with this model class.  

\section{Model evaluation}
We here present the data and objective function used for our analyses. We explain individual learning approaches for the different model classes and present the results of our analysis including individual modeling examples. 

\subsection{Data} 
We predict individual time series of platelet counts of 360 selected patients of the NHL-B trial \cite{pfreundschuh_two-weekly_2004-1, pfreundschuh_two-weekly_2004}. All patients had given informed consent and studies were approved by responsible ethics committees and were carried out in accordance with the principles of good clinical practice and the declaration of Helsinki. Details on ethics committees and reference numbers can be found in the respective publications of the studies used. Patients received CHOP-like treatment for high-grade non-Hodgkin's lymphoma. We select patients with platelet data available for at least four therapy cycles and more then one blood count per cycle. Patients are grouped according to cycle duration (14 vs. 21 days). Moreover, we consider different scenarios of data availability, namely sparse (Sp, N=225) vs. dense (De, N=135) time series. As such, we determine the following four groups De14 (N=62), De21 (N=73), Sp14 (N=96) and Sp21 (N=129). An overview table can be found in the supplement. Patients receiving the d14 treatment often have more strongly fluctuating dynamics in comparison to patients receiving the d21 treatment \cite{kheifetz_modeling_2019}. Also, higher toxicities and stronger fluctuating dynamics are observed in the De group, so that these patient dynamics are typically more difficult to predict. 

Model training is performed using data from one to five therapy cycles per patient, with the objective to predict subsequent treatment cycles. We evaluate prediction performance with our SMSE measure, introduced below. 

\subsection{Individual learning approaches}
For learning purposes, we employ log-transformed platelet counts. We estimate individual parameters of the mechanistic models by likelihood  optimization, while applying a penalty for substantial deviations from the population mean. This regularization balances individual model fidelity with population consistency, avoiding deviations from physiologically plausible parameter ranges. Details are provided in the supplement. We implement the mechanistic models in Python \cite{python_software_foundation_python_2020}, details can be found in the provided code. All patients and data availability scenarios are processed in parallel on a CPU cluster due to time constraints. Parallelization is implemented using SLURM array jobs. The cluster is equipped with Intel Xeon Platinum 8470 cores @ 2.00 GHz. A model for a single patient and data availability scenario can be fit on a local machine in reasonable time. For a single patient, we employ two cores with four threads in total, and need about 120 MB of RAM and up to six minutes of compute time. In total, the estimation of individual parameters for all patients with the mechanistic models takes 182 core hours execution time. 

To learn individual dynamics with the hybrid UDE models, we first estimate parameters of the Friberg model by fitting its predictions to the individual patient data. For the UDE-add model, we directly use these parameters in the UDE and train the network part in a second step including the individual mechanistic parameters for further improvements. In contrast, for the UDE-rep model we cannot directly use the Friberg parameters. Therefore, we employ a transfer learning approach in the second step by training the network with the solution of the Friberg model for the specific patient, i.e. we use synthetic data to learn the model. Finally, we fine-tune the network model and the mechanistic parameter settings based on the original patient data. Details of the learning process such as final employed learning rate, optimization algorithm etc. can be found in the provided code. We supply hyperparameter search ranges in the supplement. We also implement the hybrid models in Julia \cite{bezanson_julia_2017} due to package availability and use the same CPU cluster as for the mechanistic models. For both hybrid models, per patient and data availability scenario, we employ two cores with four threads and need around two GB of RAM and 15-20 minutes of training time. In a few extreme cases we need up to three hours of training time. To train all UDE-add models, we need 1.4k core hours, and for all UDE-rep models 3.1k core hours. For hyperparameter optimization and including failed attempts, our experiments need 101k core hours in total.

For the ARX-GRU model, we employ a transfer learning approach to cope with the sparse data problem of clinical data. Similar as for the hybrid models, we first parameterize the Friberg model for an individual patient. Because the ARX-GRU approach is purely data-driven, we simulate multiple virtual therapy scenarios \cite{steinacker_individual_2023} for pre-training the model. Afterwards, the model is fine-tuned on the real patient data. Details of this process are given in the provided code. We implement the ARX-GRU model in Python  and also train models in parallel on the mentioned CPU cluster for time constraints. Parallelization is implemented using Ray v1.13 \cite{moritz_ray_2017}, therefore a time estimate for a single patient cannot be made, although it is possible to train the model for a single patient and data availability scenario on local resources. To train all patients in parallel we use one compute node with 64 cores for 23 hours in total and we need up to 130 GB of RAM. We specify that two cores are reserved at a time per patient and data availability scenario. The full computation to train the individual ARX-GRU models requires 1.5k hours of CPU time. Further details can be found in the provided code.

\subsection{Objective function}
The objective function we use to train the models and to compare the performance is a modified mean-squared error (SMSE). More precisely, we compare the model predictions at time points $t$, $t-1$ and $t+1$ with the data point at $t$ using a weighted average. A justification is that exact time points of platelet measurements are often not available in clinical records. This modified objective function proved to be advantageous for training and accessing performance of individual models for sparse individual data in the past \cite{steinacker_predicting_2024}. The formula for the SMSE reads as follows: 
\begin{equation} \label{eqn:SMSE}
    \text{SMSE} = \frac{1}{N}\left[ \sum_i (y_i - \hat{y}_i)^2 + 0.3\sum_i \left((y_i - \hat{y}_{i-1})^2 + (y_i - \hat{y}_{i+1})^2 \right)\right],
\end{equation}
where $y_i$ represents the measured log-transformed platelet count and $\hat{y}_i$ represents the respective model prediction. The total number of data points evaluated is denoted by $N$. The exact implementation of the SMSE can be found in the provided code linked in the supplement.

\subsection{Comparison of prediction performances}
By our study, we want to find the optimal approach of combining mechanistic and data driven approaches to describe and predict individual time courses of platelet dynamics under chemotherapy. While expecting that there might be no unique modeling / learning approach which performs best in every scenario, we consider different data situations, i.e. we consider differently strong dynamics by comparing 14d vs. 21d schedules and data sets of dense vs. sparse individual data. For the different scenarios, we explore whether the data-driven ARX-GRU framework outperforms the mechanistic models (Friberg, Henrich, Mangas-Sanjuan (MS) and the revised Mangas-Sanjuan model (MS-rev). We also analyze whether the mechanistic models can be improved by hybrid modeling using UDE-rep or UDE-add. For all approaches and scenarios, we consider different numbers of therapy cycles used for training and evaluate the prediction performances at the subsequent cycles not used in training in terms of the average SMSE in the respective group. 

\begin{figure}
    \centering
    \includegraphics[width=0.98\textwidth]{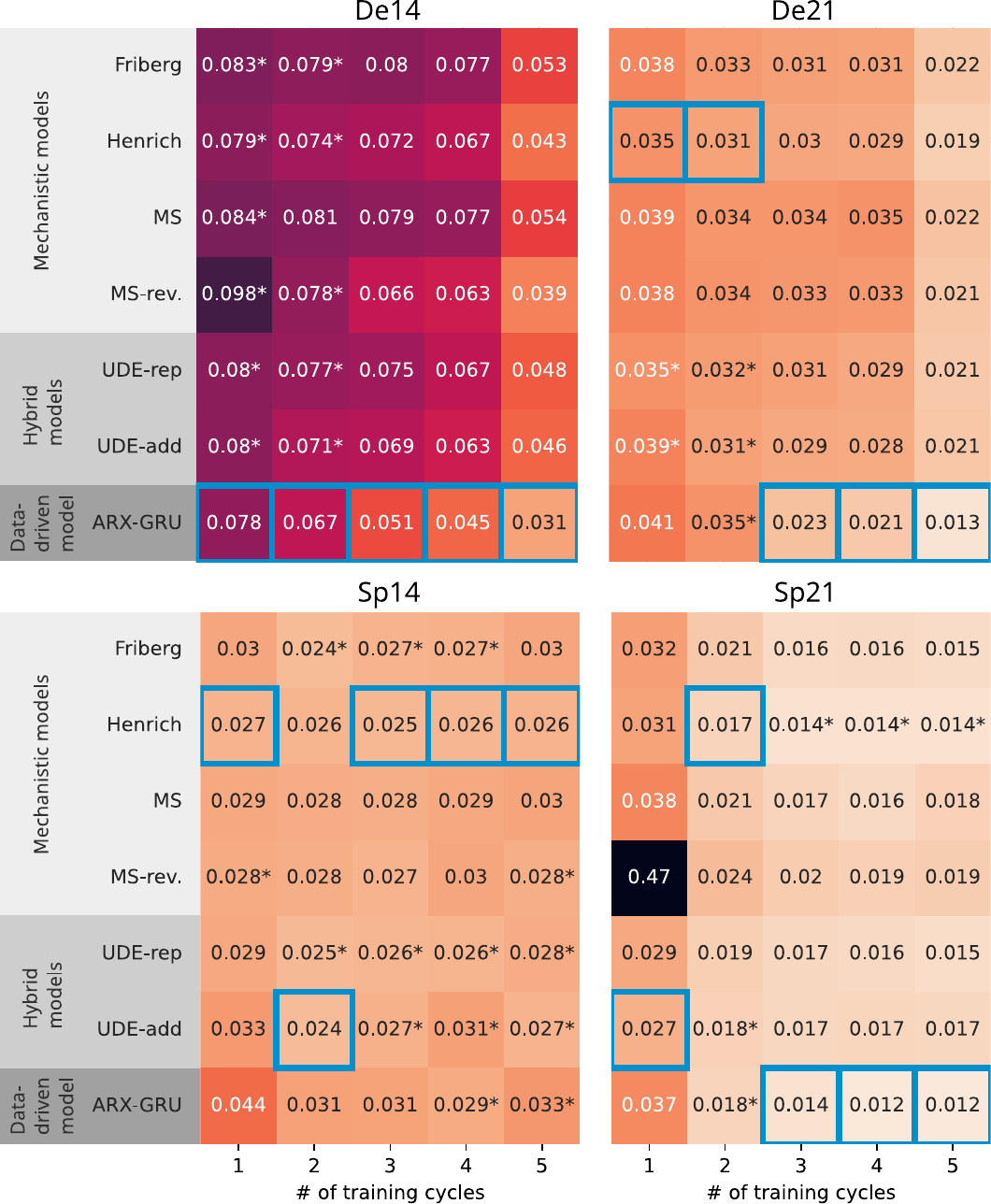}
    \caption{Average prediction performance per patient group and data scenario. We compare performances of the different modeling approaches (rows) using SMSE (see \Cref{eqn:SMSE}) averaged over all patients of the respective groups (panels) per number of therapy cycles used for training (columns). The SMSE is calculated for each patient using data and corresponding model predictions from all  cycles not used for model training. The average of SMSE across all patients is calculated and compared between groups. To emphasize differences in SMSE across modeling approaches, we use a color scale where lighter colors represent lower SMSE values, indicating better prediction performance. The best model per column is indicated by a blue box and SMSE values not significantly inferior compared to the best performance per column are marked by an asterisk.}
    \label{fig:hm}
\end{figure}

Results are presented in \Cref{fig:hm}. The four data scenarios (dense vs. sparse, 14 vs. 21 days cycle duration) are presented as heatmap panels. Compared models are represented by rows while number of therapy cycles used for training are represented by columns. The best results per column are marked by a blue box. Model results which are not significantly inferior (one-sided Wilcoxon test, $p=0.05$) compared to the best performing model of the given scenario are marked by an asterisk. 

For the De14 group the ARX-GRU approach outperforms all other modeling approaches, which becomes significant after three training cycles. For the De21 group, ARX-GRU outperforms all other models if three or more cycles are available. For two cycles or less, the Henrich model is superior, although the difference to the ARX-GRU is significant if only one training cycle is provided. The UDE frameworks perform comparably to the Henrich model.

For the Sp14 group, the Henrich model was superior except for two training cycles where the UDE-add was significantly better. In most situations, the Henrich model and the UDE frameworks did not differ significantly. For the Sp21 group the situation was similar as for the De21 group, i.e. ARX-GRU was superior after three training cycles but the difference to the Henrich model was not significant. For one training cycle, UDE-add and for two training cycles the Henrich model performed best. The bad performance of the MS-rev was due to one outlier with particularly bad fitting behavior.

\subsection{Examples}

To show the individual strengths and weaknesses of each approach and to illustrate the learning process of the models, we present individual time courses and selected model predictions in \Cref{fig:pats}. Results of all patients and models are provided in the supplement.

\begin{figure}[h]
   \centering
   \includegraphics[width=\textwidth]{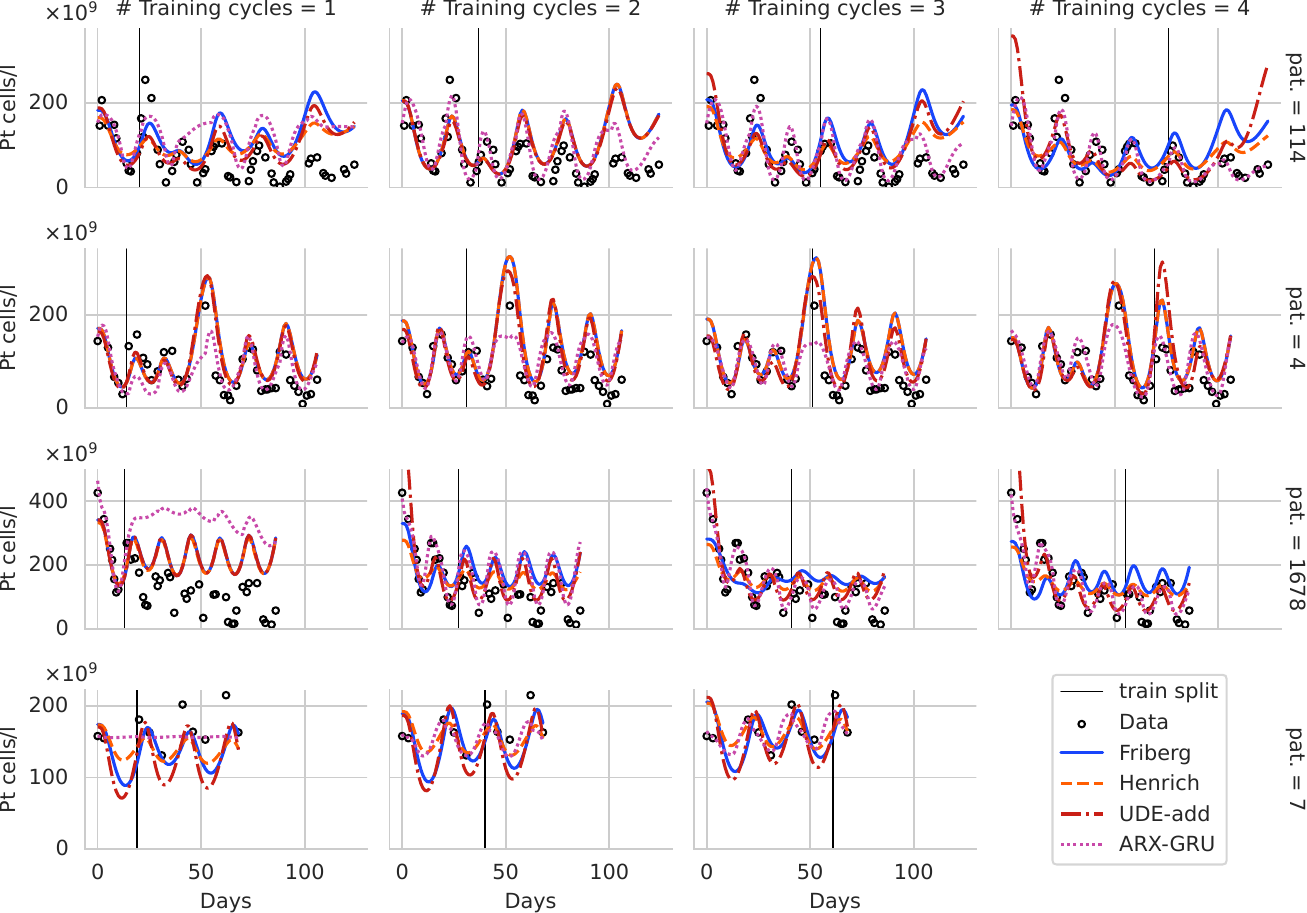}
   \caption{Individual predictions for four selected patients. We present predictions of the Friberg (blue), Henrich(orange, dashed), ARX-GRU (red, dash-dotted) and UDE-add (pink, dotted) model for NHL patients 4, 7, 114 and 1678. The split between training and prediction time points is indicated by a vertical black line. Data points are shown as black circles. For improved legibility, we use different scales per patient (rows). While patients 4, 114 and 1678 are from group De14, patient 7 belongs to group Sp21.}
    \label{fig:pats}
\end{figure}

Patient \#114 belongs to group De14 and expressed strong toxicity in all treatment cycles. Using only one therapy cycle for training resulted in poor predictions of the nadir phases by all examined models. For two or more cycles provided, the ARX-GRU model predicts nadir phases well. Given four cycles of data, the UDE-add model also matches nadir phases, although the last nadir phase is poorly matched, as was the case for all mechanistic models. The Friberg model and the Henrich model could not improve with more data provided. This shows the flexibility of data-driven approaches in comparison to purely mechanistic approaches in adapting to individual time courses. We conclude that in the case of strong toxicity, the ARX-GRU model outperforms the other models, given data of at least two cycles and not too sparse time series.

Patient \#4 also belongs to group De14 and showed irregular dynamics due to a treatment delay prior to the fourth cycle resulting in a sharp increase in platelet counts. The ARX-GRU model could not predict this spike well, possibly due to strong regularization constraints. However, the mechanistic models including the UDE-add model predicted the spike. For this patient, the predictions of the ARX-GRU match observed nadirs best, followed by the UDE-add model. In contrast to the Friberg and Henrich model, the ARX-GRU predictions of the nadir phases improve with more data provided. This illustrates the potential of combining mechanistic and data-driven modeling approaches, i.e. biologically expected behavior and model flexibility.

Patient \#1678 from group De14 shows a strong cumulative toxicity over therapy cycles with a fast decline starting from a higher then average initial cell count. With only one therapy cycle of data provided, all examined models over-predict nadir phases of subsequent cycles substantially. If more data are provided, the ARX-GRU and the UDE-add model both improve their nadir predictions, while the Friberg model and the Henrich model do not improve relevantly. This is somewhat  surprising, as the Henrich model was supposed to better predict cumulative toxicity. 

Patient \#7 belongs to group Sp21. For this patient, only two observations per cycle are available and we presume that the nadir phases are not captured well by the data. Due to strong regularization penalties, the ARX-GRU predicts a near constant if only one cycle of data is provided. The mechanistic models and the UDE-add model in part represent an average mechanistic population dynamic in this situation. This demonstrates the advantage of incorporating mechanistic knowledge in case of sparse data. For two cycles or more, all models agree well with the observed data. We conclude, that data quality and quantity is crucial to employ models as the ARX-GRU and that for very sparse data, a mechanistic or hybrid approach can be the better choice. 

\section{Conclusion}
In this work we developed hybrid and data-driven approaches for modeling individual time series of platelets under chemotherapy and compared their prediction performance with those of popular mechanistic models. We observed a strong advantage of the data-driven ARX-GRU model for individual treatment planning, in particular for patients at high toxicity risk. It outperformed all other approaches for the groups De14 and De21 showing deeper nadir phases of platelet counts. However, a sufficient data base is required - at least three treatment cycles with at least 3 measurements per cycle. For cases of sparser data or with less cycles available, mechanistic or hybrid models are advantageous. However, one needs to acknowledge that with these sparser data, the important nadir phase of cell counts might not be captured by the data, and with it, the prediction models do not reflect the true underlying course of the patient. In our study, the hybrid models, UDE-add and UDE-rep outperformed the simple Friberg model and showed comparable performance to refined mechanistic models but without additional biological assumptions.

\paragraph{Impact.} In this work, we considered dynamics of platelet counts only. We believe that our methodology can be generalized to other hematologic lineages, which we aim to investigate in the future including multi-lineage approaches. Additionally, our approach allows to include further therapeutic interventions such as hematopoietic growth factors often administered in high-risk cases or therapies with high cytotoxicity. Thus, our models could significantly enhance the accuracy of predicting the impact of chemotherapy on a patient's hematopoietic system, thereby assisting medical professionals in developing individualized treatment plans. In doing so, our methods can help to optimize treatment outcomes and improve a patients' quality of life. Finally, we present a framework of assessing the reliability of model predictions helping to select the most appropriate model in dependence on the available data and the anticipated toxicity risk.

\paragraph{Limitations and future work.} Reliability of all model predictions depend heavily on the quality of the data of the individual patient. The coverage of nadir phases is crucial for accurate learning of individual dynamics, as well as assessing model performance. For the data-driven ARX-GRU model, we find that data of at least two treatment cycles is needed for successful learning. Per cycle, the initial decline, the nadir phase and the recovery of cell counts should be covered by measurements.

Our hybrid models UDE-add and UDE-rep are UDE extensions of the mechanistic Friberg model, and thus, largely constrained by this model structure serving as the basis. While this generally preserves the explainability of predictions, their solutions could remain close to those of the Friberg model, in particular in case of sparse data.
We deliberately focused on two hybrid models representing only minor adjustments of the Friberg model. It is conceivable that other or more comprehensive network components could  improve flexibility and would allow more precise predictions of individual patient dynamics. In future work, we plan to explore different hybrid architectures in this regard, drawing inspiration from the mechanistic refinements of the Friberg model examined in this study. 

\begin{ack}
The authors acknowledge the financial support by the Federal Ministry of Education and Research of Germany and by Sächsische Staatsministerium für Wissenschaft, Kultur und Tourismus in the programme Center of Excellence for AI-research „Center for Scalable Data Analytics and Artificial Intelligence Dresden/Leipzig“, project identification number: ScaDS.AI. 

This project was funded by “ChemoTox-AI”, a project financed by the German Federal Ministry of Education and Research (BMBF) within the Computational Life Science line of funding (BMBF 031L0261). 

Yuri Kheifetz was funded by "TOPIC", a project financed by the German Federal Ministry of Education and Research (BMBF) (BMBF 13GW0760B). 

The authors gratefully acknowledge the computing time made available to them on the high-performance computer at the Technical University Dresden (NHR center). This center is jointly supported by the Federal Ministry of Education and Research and the state governments participating in the NHR network (www.nhr-verein.de/unsere-partner).

\end{ack}

%%%%%%%%%%%%%%%%%%%%%%%%%%%%%%%%%%%%%%%%%%%%%%%%%%%%%%%%%%%%

\bibliography{main}

%%%%%%%%%%%%%%%%%%%%%%%%%%%%%%%%%%%%%%%%%%%%%%%%%%%%%%%%%%%%

\appendix

\section{Technical Appendices and Supplementary Material}
The following supplemental materials are provided to support the findings discussed in the main manuscript:

\paragraph{Code availability} We provide the full code for our models at \url{https://github.com/earlgreymatchalatte/Hybrid_Data-Driven_Models}.

\paragraph{Hyperparameter details} All search ranges for the hyperparameters explored are included in \texttt{supplement\_1.pdf}, Section A.

\paragraph{Mechanistic models} Differential equations for all mechanistic models, population-average parameters, and the objective function are provided in \texttt{supplement\_1.pdf}, Section B.

\paragraph{Data description} A tabular overview of the dataset used in model comparison, including group details, is given in \texttt{supplement\_1.pdf}, Section C.

\paragraph{Individual predictions} Predictions for all models, patients, and data availability scenarios are shown in \texttt{supplement\_2.pdf}.

\end{document}

\appendix

\section{Hyperparameter Details}
We performed hyperparameter optimization for the hybrid models UDE-rep and UDE-add. \Cref{tab:hp} summarizes all optimized hyperparameters along with their corresponding search ranges.\\

Due to the large number of hyperparameters in UDE-rep, we divided the optimization into two stages. In the first stage, we tuned the model architecture and training regimen during pre-training, using virtual patient time series generated by the Friberg model \cite{friberg_model_2002} with population-average parameters, see \Cref{sec:friberg}. This step was conducted entirely on synthetic data to reduce computational costs. At the second stage, we fine-tuned the remaining hyperparameters using real patient data, specifically a randomly selected half of the De group. For UDE-add, we optimized all hyperparameters directly on the same half of the De group. In contrast, the data-driven ARX-GRU model uses the hyperparameters reported in the original publication \cite{steinacker_predicting_2024}.

\begin{longtable}{ >{\raggedright\arraybackslash}m{2cm} >{\raggedright\arraybackslash}p{2.8cm} >{\raggedright\arraybackslash}p{2.5cm} >{\raggedright\arraybackslash}m{3.2cm} p{4.5cm}}
\caption{Detailed listing of hyperparameters used in the unified differential equation (UDE) modeling framework, split by model and training stage. Each row specifies the parameter name, search range, relevant code configuration (where applicable), and a brief functional description.} \label{tab:hp}\\
\toprule \rowcolor{gray!10}
Model/ Training stage &
  Hyper\-parameter &
  Search range &
  Code reference$\quad$ (if applicable) &
  Description \\
  \midrule
  \midrule
  \endhead
  \cellcolor{white} 
\shortstack{UDE-rep \\ pre-training} &
  Number of hidden layers &
  {[}1,2,3{]} &
   &
  Number of hidden layers in the network model part. \\ \altrowcolor \cellcolor{white} 
 &
  Number of nodes per layer &
  {[}3, 5, 10, 20, 50{]} &
   &
  Number of neurons in each hidden layer. \\ \cellcolor{white} 
 &
  Activation function &
  {[}sigmoid, tanh, rbf, linear{]} &
   &
  Activation function used in hidden layers. \\ \altrowcolor \cellcolor{white} 
 &
  l2 Regularization &
  {[}0, 0.1, 1, 10, 100{]} &
  \texttt{l2reg} &
  L2 regularization strength for weight decay; helps to prevent overfitting. \\\cellcolor{white} 
 &
  Optimization algorithm &
  {[}Adam, Sophia, LBFGS, BFGS{]} &
   &
  Optimization algorithm, also includes combinations applied sequentially. \\  \altrowcolor\cellcolor{white} 
 &
  Initial learning rate &
  {[}1e-4, \hspace{0pt}\allowbreak1e-3, \hspace{6pt}\allowbreak5e-3, \hspace{0pt}\allowbreak1e-2{]} &
  \texttt{lr\_start} &
  Learning rate for Adam optimization algorithm \\  \cellcolor{white} 
 &
  Learning rate decay &
 {[}-0.9, -0.7, -0.5, -0.3, -0.1, -0.05, 0{]} &
  \texttt{decay} &
  Exponential factor applied to learning rate at decay step. \\ \altrowcolor \cellcolor{white} 
 &
  Decay step length &
  {[}200, 400, 600{]} &
  \texttt{decay\_steps} &
  Number of epochs before next learning rate adjustment. \\  \cellcolor{white} 
 &
  Decay steps &
  {[}3,5,7,9{]} &
  \texttt{steps} &
  Total number of decay steps applied during training. \\ \altrowcolor \cellcolor{white} 
 &
  Coupling factor steady state &
  {[}0, 0.1, 1{]} &
  \texttt{couple} &
  Coupling factor of steady state deviation penalty. \\  \cellcolor{white} 
 &
  Coupling factor follow up &
  {[}0, 0.1, 1{]} &
  \texttt{s} &
  Coupling factor of steady state deviation penalty applied to follow up prediction. \\ \altrowcolor \cellcolor{white} 
 &
  Coupling factor SMSE &
  {[}0, 0.1, 1{]} &
  \texttt{step\_s, step\_opt} &
  Coupling factor for influence of neighboring day predictions in SMSE. \\  \cellcolor{white} 
 &
  Friberg parameter scaling &
  {[}0.1, 0.2,0.5, 1.{]} &
   &
  Spread parameter (in percent) for standardization relative to population parameters. This is performed to achieve better optimization together with network parameters. \\  
  \midrule \altrowcolor \cellcolor{rowgray} 
  \shortstack{UDE-rep\\ fine-tuning} 
  &
  Skip pre-training &
  {[}True, False{]} & \hspace{3.2cm}
  \texttt{use\_pop\_weights}, \texttt{use\_pop\_params}, \texttt{use\_preTL\_weights} &
  If True, skips pre-training and starts directly with the fine-tuning stage.
   \\  \cellcolor{rowgray} 
 &
  Optimization Algorithm &
  {[}Adam, Sophia, LBFGS{]} &
  \hspace{2cm} \texttt{skip\_adam}, \texttt{use\_sophia} &
  Optimization algorithm, also includes combinations applied sequentially. \\ \altrowcolor\cellcolor{rowgray} 

 &
  Learning rate &
  {[}1e-4,1e-3, 5e-3, 1e-2{]} &
  \texttt{lr\_startTL} &
  Learning rate for Adam optimization algorithm. \\ \cellcolor{rowgray} 

  &
  Learning rate decay &
  {[}-0.9, -0.7, -0.5, -0.3, -0.1, -0.05, 0{]} &
  \texttt{decayTL} &
  Exponential factor applied to learning rate at decay step. \\\altrowcolor \cellcolor{rowgray} 
 &
  l2 Regularization &
  {[}0, 0.1, 1, 10, 100{]} &
  \texttt{l2regTL} &
  L2 regularization strength for weight decay; helps to prevent overfitting. \\ 
  \midrule  \cellcolor{white} 
 UDE\_add & \rownum=1\relax 
  Number of hidden layers &
  {[}1,2,3{]} &
  \texttt{config} &
  Number of hidden layers in the network model part. \\ \altrowcolor \cellcolor{white}
 &
  Number of nodes &
  {[}3, 5, 10, 20, 50{]} &
  \texttt{config} &
  Number of neurons in each hidden layer. \\  \cellcolor{white}
 &
  Activation function &
  {[}sigmoid, tanh, rbf, linear{]} &
   &
  Activation function used in hidden layers. \\ \altrowcolor \cellcolor{white}
 &
  l2 Regularization &
  {[}0, 0.1, 1, 10, 100{]} &
  \texttt{l2reg} &
  L2 regularization strength for weight decay; helps to prevent overfitting. \\ \cellcolor{white}
 &
  Optimization algorithm &
  {[}Adam, Sophia, LBFGS{]} &
   &
  Optimization algorithm, also includes combinations applied sequentially. \\ \altrowcolor \cellcolor{white}
 &
  Initial learning rate &
  {[}1e-4,1e-3, 5e-3, 1e-2{]} &
  \texttt{lr\_start} &
  Learning rate for Adam optimization algorithm. \\\cellcolor{white}
 &
  Learning rate decay &
  {[}-0.9, -0.7, -0.5, -0.3, -0.1, -0.05, 0{]} &
  \texttt{decay} &
  Exponential factor applied to learning rate at decay step. \\ \altrowcolor \cellcolor{white}
 &
  Decay step length &
  {[}200, 400, 600{]} &
  \texttt{decay\_steps} &
  Number of epochs before next learning rate adjustment. \\\cellcolor{white}
 &
  Decay steps &
  {[}3,5,7,9{]} &
  \texttt{steps} &
  Total number of decay steps applied during training. \\ \altrowcolor \cellcolor{white}
 &
  Coupling factor steady state &
  {[}0, 0.1, 1{]} &
  \texttt{s} &
  Coupling factor of steady state deviation penalty. \strut \\\cellcolor{white}
 &
  Coupling factor SMSE &
  {[}0, 0.1, 1{]} &
  \texttt{step\_s} &
  Coupling factor of influence of neighboring day predictions in SMSE. \\ \altrowcolor \cellcolor{white}
 &
  Friberg parameter scaling &
  {[}0.1, 0.2,0.5, 1.{]} &
   &
  Spread parameter (in percent) for standardization relative to population parameters. This is performed to achieve better optimization together with network parameters.\\
  \bottomrule
\end{longtable}

\section{Mechanistic models}
In this work, using the term mechanistic model refers to a bio-mathematical differential equations model, where the model equations are intended to describe known biological processes or treatment effects. This is also known as pharmacokinetic/pharmacodynamic (PK/PD) modeling. In the following, we present the model equations not introduced in the main part. For simulations, we used average population parameters. We also present the  objective function employed to estimate individual-specific parameters.

\subsection{Objective function}
The objective function $\mathcal{L}$ to estimate individual model parameters $p_j$ reads as follows
\begin{align}
    \mathcal{L} &= \mathcal{L}_{data} + \mathcal{L}_{param},\\
                &= \frac{1}{N} \sum_{i=1}^N (y_i - \hat{y}_i)^2 + \sum_j \frac{(\ln(p_j) - \ln(p_j^0))^2}{\sigma^2},
\end{align}
with data $y_i$ for time point $i$ and corresponding model prediction $\hat{y}_i$, as well as population average parameters $p_j^0$ and spread constraint $\sigma$ penalizing the deviation from the population average. We set this constraint parameter to $5$. This objective function balances two aspects: the data-fitting term $\mathcal{L}_{data}$ and a regularization term $\mathcal{L}_{param}$, which incorporates population-derived parameter constraints. This regularization aims at preventing overfitting by discouraging individual parameter estimates deviating strongly from the population average.

\subsection{Friberg model} \label{sec:friberg}
We provide the differential equations of the model of \citeauthor{friberg_model_2002} \cite{friberg_model_2002} in the main text, section 2. We assume the following population averages, proposed in the literature: feedback parameter $\gamma = 0.316$, transit time $\text{MTT}=\SI{195}{\hour}$ \cite{joerger_population_2007} and steady state $C_0 = \SI{270e9}{cells\per \liter}$ \cite{warny_arterial_2019}. The transition rate $k_{tr}$ is derived by $k_{tr} = 4/\text{MTT}$. For the linear drug effect function, we assume
\begin{equation}
E_{drug}(t) =
\begin{cases}
E_{eff}, & \text{if } t \in T ,\\
0, & \text{otherwise},
\end{cases}
\end{equation}
where $T$ is the set of treatment days. We assume the population average $E_{eff} = 2$ as the result of population analysis.

\subsection{Henrich model}
The model of \citeauthor{henrich_semimechanistic_2017}, \cite{henrich_semimechanistic_2017} is a refinement of the Friberg model and includes slowly replicating pluripotent stem cells in the bone marrow as an additional compartment. This refinement is intended to describe cumulative toxicity, which is the long term toxic effect of multicyclic chemotherapy on the hematopoietic system. The model's  ordinary differential equations read as follows:
\begin{align}\label{eqn:henrich}
    \frac{d}{dt}S &=k_{s}  (1- E_{drug})  \left(\frac{C_0}{C}\right)^\gamma  S  - k_{s}  S \\
    \frac{d}{dt}P &= k_{p}  P  (1-E_{drug}) \left(\frac{C_0}{C}\right)^\gamma - k_{tr}  P + k_{s}  S\\
    \frac{d}{dt}T_1 &= k_{tr}  P - k_{tr}  T_1\\
    \frac{d}{dt}T_2 &= k_{tr}  T_1 - k_{tr}  T_2\\
    \frac{d}{dt}T_3 &= k_{tr}  T_2 - k_{tr}  T_3\\
    \frac{d}{dt}C &= k_{tr}  T_3 - k_{c}  C
\end{align}

Assuming steady state without therapy, $k_{p}$ is estimated as fraction $f_{tr}$ of the transition rate constant.
\begin{align}
    k_{p} &= f_{tr}  k_{tr}\\
    k_{s} &= (1- f_{tr})  k_{tr}
\end{align}
Setting $f_{tr} = 1$ results in the original Friberg model. The population average parameters are identical to that of the Friberg model, and the fraction constant $f_{tr} = 0.7$ is assumed based on population analysis.

\subsection{Mangas-Sanjuan model}

The model of \citeauthor{mangas-sanjuan_semimechanistic_2015}, \cite{mangas-sanjuan_semimechanistic_2015} is a another refinement of the Friberg model, which incorporates both proliferative and quiescent stem cell populations. In comparison to the Friberg model, two additional compartments $Q_1$  and $Q_2$  for quiescent cells are introduced, as well as an additional fraction rate constant $f_p$. Again, the refinement is intended to describe cumulative toxicity:

\begin{align}\label{eqn:ms}
    \frac{d}{dt}P &= k_{p}  P  (1-E_{drug}) \left(\frac{C_0}{C}\right)^\gamma - k_{tr}  f_p  P + k_{cyc}  (Q_2 - (1-f_p)  P) \\
    \frac{d}{dt}Q_1 &= k_{cyc}  ((1 - F_P)  P - Q_1) \\
    \frac{d}{dt}Q_2 &= k_{cyc}  (Q_1 - Q_2) \\
    \frac{d}{dt}T_1 &= k_{tr}  F_P  P - k_{tr}  T_1\\
    \frac{d}{dt}T_2 &= k_{tr}  T_1 - k_{tr}  T_2\\
    \frac{d}{dt}T_3 &= k_{tr}  T_2 - k_{tr}  T_3\\
    \frac{d}{dt}C &= k_{tr}  T_3 - k_{c}  C
\end{align}

For steady state without therapy, $k_{p}$ is estimated as fraction $f_p$ of the transition rate constant.
\begin{align}
    k_p &= f_p  k_{tr}\\
    P_0 &= \frac{C_0}{f_p} \\
    Q_{1,0} &= Q_{2,0} = (1-f_p)  P_0  %steady state assumption
\end{align}
Setting $f_p = 1$ and $k_{cyc} =0 $ results in the original Friberg model. The population average parameters are again identical to that of the Friberg model. We employ the additional average population parameter estimates $f_p = 0.58$ and $k_{cyc} = \SI{1.9}{\per \day}$ from the original publication \cite{mangas-sanjuan_semimechanistic_2015}.

\subsection{Revised Mangas-Sanjuan model}
Studies suggest that hematopoiesis operates as a stochastic rather than deterministic process \cite{abkowitz_evidence_1996}. The exact ratio of dormant (quiescent) to active hematopoietic stem cells (HSCs) remains uncertain. However, murine studies indicate that the majority of HSCs reside in a dormant state and are activated primarily in response to injury or stress \cite{wilson_hematopoietic_2008, busch_fundamental_2015}. In humans, \citeauthor{catlin_replication_2011} \cite{catlin_replication_2011} reported a very low turnover rate, estimating that HSC division occurs approximately once every 25–50 weeks, supporting the existence of a relatively large quiescent stem cell population. To capture this behavior in our model, we introduce an additional rate constant, $k_{cyc2}$, governing transitions from the second quiescent compartment to proliferation. To reflect the low turnover rate, $k_{cyc2}$ is assumed to be significantly smaller than the original $k_{cyc}$. This revised Mangas-Sanjuan model is described by the following set of equations and parameters:

\begin{align}\label{eqn:msr}
    \frac{d}{dt}P &= k_{P} P (1-E_{drug}) \left(\frac{C_0}{C}\right)^\gamma - k_{tr} f_{P}P + k_{cyc2}\, Q_2 - k_{cyc} (1-f_{P})  P), \\
    \frac{d}{dt}Q_1 &= k_{cyc}  ((1 - f_P)  P - Q_1), \\
    \frac{d}{dt}Q_2 &= k_{cyc}\,  Q_1 - k_{cyc2}\, Q_2, \\
    \frac{d}{dt}T_1 &= k_{tr}  f_P  P - k_{tr}  T_1,\\
    \frac{d}{dt}T_2 &= k_{tr}  T_1 - k_{tr}  T_2,\\
    \frac{d}{dt}T_3 &= k_{tr}  T_2 - k_{tr}  T_3,\\
    \frac{d}{dt}C &= k_{tr}  T_3 - k_{c}  C,
\end{align}
with the steady state assumption 
\begin{align}
    \frac{k_{cyc}}{k_{cyc2}} Q_{1,0} = Q_{2,0}.
\end{align}
The population parameter estimates are identical to that of the original Mangas-Sanjuan model, with the exception of the additional parameter $k_{cyc2} = k_{cyc}/60$ which we obtained by population analysis. The drug effect estimate had to be adjusted accordingly $E_{eff} = 120$.

\section{Description of data}

We consider individual time series of patients treated in the framework of the NHL-B study \cite{pfreundschuh_two-weekly_2004-1, pfreundschuh_two-weekly_2004}. Patients were treated with CHOP-like therapies for high-grade non-Hodgkin's lymphoma. We consider patients who , received 4-6$\times$ CHOP-14/21 or CHOEP-14/21 and exclude patients who received platelet transfusions. In our analysis, we group patients regarding sparsity of the recorded individual time series and regarding treatment regiments to access the impact of these factors on prediction and model performances. This resulted in four patient groups; De14, De21, Sp14 and Sp21. We present group features in \Cref{tab:pat}.

\begin{table}[h]
\centering
\caption{We present distributions of risk factors and treatment for all considered patients, and separately, for the subgroups of dense (De) and sparse (Sp) time series data per treatment regimen (d14, d21). Additionally, we provide the distribution of recorded treatment cycles, as patients received between four and six treatment cycles in total. For risk factors age>60 years, sex and treatment as well as for recorded treatment cycles, total numbers are given. For the initial platelet count, we provide the average value per patient (sub-)group. }
\begin{tabular}{lllllllll}
\toprule
                            &          & all & De  & Sp  & De14 & De21 & Sp14 & Sp21 \\
                            \midrule
\multicolumn{2}{l}{Risk factor}        &     &     &     &      &      &      &      \\

\multirow{2}{*}{age}        & young ($\leq 60$ years )    & 207 & 59  & 148 & 25   & 34   & 64   & 84   \\
                            & old ($> 60$ years )     & 153 & 76  & 77  & 37   & 39   & 32   & 45   \\
\multirow{2}{*}{sex}        & male     & 202 & 65  & 137 & 27   & 38   & 57   & 80   \\
                            & female   & 158 & 70  & 88  & 35   & 35   & 39   & 49   \\
\multirow{4}{*}{treatment}  & CHOP-14  & 71  & 25  & 46  & 25   & -    & 46   & -    \\
                            & CHOEP-14 & 87  & 37  & 50  & 37   & -    & 50   & -    \\
                            & CHOP-21  & 94  & 28  & 66  & -    & 28   & -    & 66   \\
                            & CHOEP-21 & 108 & 45  & 63  & -    & 45   & -    & 63   \\
\multicolumn{2}{l}{Mean initial platelet count $\times 10^9$ \si{\per \litre}} & 291.3 & 276 & 300.5 & 270.3 & 280.9 & 306.7 & 295.7 \\
\midrule
\multicolumn{2}{l}{\# recorded cycles} &     &     &     &      &      &      &      \\
                            & 4        & 360 & 135 & 225 & 62   & 73   & 96   & 129  \\
                            & 5        & 351 & 128 & 223 & 56   & 72   & 95   & 128  \\
                            & 6        & 337 & 119 & 218 & 54   & 65   & 95   & 123  \\
\midrule
\multicolumn{2}{l}{total}              & 360 & 135 & 225 & 62   & 73   & 96   & 129  \\

\bottomrule
\end{tabular}
\label{tab:pat}
\end{table}

\clearpage
\bibliography{main}